# Truncated Hilbert Transform: Uniqueness and a Chebyshev Series Expansion Approach[1]

**Jiangsheng You**

Cubic Imaging LLC, 18 Windemere Dr., Andover, MA 01810

jshyou@gmail.com

**Abstract**

We derive a stronger uniqueness result if a function and its truncated Hilbert transform are known on a same interval by the Sokhotski–Plemelj formula. Using the Chebyshev series expansion, we find an explicit procedure to derive the series coefficients of a function from its truncated Hilbert transform through Lagrange interpolations, and then suggest two numerical methods to estimate the series coefficients. Last, we present computer simulation results to show that the extrapolative procedure produces good numerical results.

## I. Background and introduction

The investigation of the finite Hilbert transform (FHT) has been a research topic for a long time in history from both mathematical interests and practical needs in many applications. Some early works on the theoretical investigation and practical applications in fluid mechanic can be found in [1-7]. The relation between the line integral and the Hilbert transform was first derived by Gelfand and Graev in [8]. The introduction of using the FHT to reconstruct images from partial data in single photon computed tomography (SPECT) and computed tomography (CT), started from [9-11]. We published a short paper toward a simple inversion of the FHT in [12] for the use in the second step of [11]. Based on one explicit inversion formula of the FHT and the analytical continuation on an interval, one uniqueness result and the stability of the analytical continuation were obtained in [13]. With the breakthrough work on the cone-beam data reconstruction [14], the arguments of [13] can be extended to the 3D cone-beam (region of interest) ROI data reconstruction [15-19]. Readers should keep in mind that the image reconstruction in these works is reduced to the inversion of the FHT or the truncated Hilbert transform (THT). Therefore, the uniqueness of the inverse THT and methods to find solutions of the inverse THT become very important. Besides CT and SPECT, the FHT and THT can be useful in the context of exponential Radon transform for many other applications such as positron emission tomography (PET), nuclear magnetic resonance (NMR) imaging, ultrasonic tomography and Doppler tomography as discussed in [20]. The projection onto convex sets (POCS) was used in the numerical experiment in [13]. Recently the authors of [18, 21, 22] investigated the single value decomposition (SVD) methods to find solutions of the inverse THT. In this paper, we will focus on the theoretical study of the uniqueness of the THT and numerical procedures to find solutions of the inverse THT without further explaining the application background.

For a function $f(t)$ of $(-1, 1)$, the Hilbert transform $F(s)$ is defined as

$$F(s) = \frac{1}{\pi} \int_{-1}^{1} \frac{f(t)}{s-t} dt. \tag{1.0}$$

---

[1] This paper was previously submitted to Inverse Problem in 2012 and the author withdrawn the submission due to schedule conflict for revision. Recently the revised paper posted on arXiv: 2002-02073.

Throughout this paper, the integral at the singular point should be understood in the sense of Cauchy principal value and we always use the pair of $f(t)$ and $F(s)$ to stand for a function and its Hilbert transform. Certainly, for a function $f(t)$ of $(-\infty, \infty)$, $F(s)$ can be defined on $(-\infty, \infty)$, which is the conventional Hilbert transform. The FHT concerns the mapping between functions defined on an finite interval such as $(-1,1)$. The FHT has many nice properties, for example, $f(t)$ can be uniquely obtained from $F(s)$ under general conditions and the Plancherel formulas hold in the weighted $L^2$ spaces [23, 24]. In the case of the inverse THT, $F(s)$ is only known on an interval $(a,b)$ which is different from $(-1,1)$, then the goal is to find $f(t)$ using available $F(s)$ in $(a,b)$. The existence of $f(t)$ is readily available from the FHT, but the uniqueness of $f(t)$ satisfying (1.0) is not obvious. The existing uniqueness results in [13, 15-19] only concern the partial uniqueness of $f(t)$ on certain subinterval of $(-1,1)$, for example, the result of [13] is that if $f(t)$ has support on $[-1,c]$ and $F(s)$ is known on $(b,1)$, $-1<b<c<1$, then $f(t)$ can be uniquely determined in $(b<c]$. The main arguments used in [13] include the explicit inversion formula of the FHT and the analytical continuation on an interval. These arguments can be extended to study the interior problem of CT image reconstruction in [15-18]. The analytical continuation is by and large a mathematical procedure and lacks a numerical procedure to be realized. These uniqueness results are not applicable to the data settings assumed in [21, 22] to find $f(t)$ in its entire support interval.

In this paper, using the Sokhotski–Plemelj formulas, we derive a stronger uniqueness result for $f(t)$ on its support interval $(-1,1)$ weaker conditions than the ones found in previously mentioned works. The result in this paper confirms that the solution by the SVD schema of [21, 22] can be unique under the assumptions in [13, 15-18]. Moreover, our uniqueness results are valid to the exponential Radon transform. We also describe one example of using such stronger uniqueness result to resolve the typical data truncation problem in tomographic imaging applications when the object is out of the field of view. We continue to explore the Chebyshev polynomials to evaluate the Hilbert transform and its inversion mentioned in [5, 25] and further studied in our recent work [24]. Surprisingly an explicit inversion can be achieved through Lagrange interpolations if the functions only include finite Chebyshev polynomials. Using the explicit Chebyshev polynomials can avoid the estimate of the eigenvalues and eigenfunctions needed in the SVD schema of [21, 22]. Numerically, the evaluation of the Chebyshev polynomials can be implemented by fast sine and cosine transformations. To numerically find a solution to the THT problem, we express $f(t)$ and $F(s)$ in the finite Chebyshev polynomial series, and then try to estimate the coefficients using the available data from both $f(t)$ and $F(s)$. Two methods are proposed to estimate the coefficients, one is to minimize a cost function and the other is an extrapolation procedure. Finally, we present computer simulation results to show the feasibility of the proposed method for practical applications.

## II.　Uniqueness of THT and cosh-weighted Hilbert transform

For a function with compact support in $R^1$, through scaling and shift, we can always make it have the support on $[-1,1]$. Without loss of generality in this paper, we assume that $f(t)$ belongs to $L^p(-1,1)$ which is defined as

$$L^p(-1,1) = \{ f(t) : \int_{-1}^{1} |f(t)|^p \, dt < \infty \}, \ p \geq 1. \tag{2.1}$$

When $p=2$, $L^2(-1,1)$ is a Hilbert space. We introduce a closed subspace $L_d^2(-1,1)$ of $L^2(-1,1)$ as follows:

$$L_d^2(-1,1) = \{f(t) : \int_{-1}^{1} \frac{|f(t)|^2}{\sqrt{1-t^2}} dt < \infty\} \tag{2.2}$$

with the inner product

$$<f,g>_d = \frac{1}{\pi} \int_{-1}^{1} \frac{f(t)g^*(t)}{\sqrt{1-t^2}} dt. \tag{2.3}$$

Hereafter, we use the superscript star for the conjugate operation of a function. Define the closed subspace $\mathcal{L}_d^2(-1,1)$ as

$$\mathcal{L}_d^2(-1,1) = \{f(t) \in L_d^2(-1,1) : <f,1>_d \equiv 0\}. \tag{2.4}$$

From [25], the Chebyshev polynomials can construct a complete base of $L_d^2(-1,1)$ and $\mathcal{L}_d^2(-1,1)$. According to [23, 24], the FHT is an isotropy from $L_d^2(-1,1)$ to $\mathcal{L}_d^2(-1,1)$. If $f(t) \in L_d^2(-1,1)$, then

$$\int_{-\infty}^{\infty} |F(s)|^2 ds = \int_{-1}^{1} |f(t)|^2 dt, \quad \int_{-1}^{1} \frac{|F(s)|^2}{\sqrt{1-s^2}} ds = \int_{-1}^{1} \frac{|f(t)|^2}{\sqrt{1-t^2}} dt. \tag{2.5}$$

We will use $C$ for the complex plane and $i = \sqrt{-1}$ for the imaginary unit. The Sokhotski–Plemelj formula is a historical result and its detailed proof can be found in many graduate text books such as [4] on the integral operator theory. In Proposition 1, we reformulate the results on the Sokhotski-Plemelj formulas under the conditions for the use in this paper.

**Proposition 1** (Sokhotski–Plemelj formulas). *Let* $f(t) \in L^p(-1,1)$, $p \geq 1$, *define*

$$\Phi(z) = \frac{1}{2\pi i} \int_{-1}^{1} \frac{f(t)}{t-z} dt. \tag{2.6}$$

Then $\Phi(z)$ is analytical in $C \setminus [-1,1]$, and for almost every $s \in (-1,1)$ and $\varepsilon > 0$, $\lim_{\varepsilon \to 0} \Phi(s+i\varepsilon)$ and $\lim_{\varepsilon \to 0} \Phi(s-i\varepsilon)$ exist and satisfy

$$\Phi^+(s) = \lim_{\varepsilon \to 0} \Phi(s+i\varepsilon) = \frac{1}{2\pi i} \int_{-1}^{1} \frac{f(t)}{t-s} dt + \frac{1}{2} f(s), \tag{2.7}$$

$$\Phi^-(s) = \lim_{\varepsilon \to 0} \Phi(s-i\varepsilon) = \frac{1}{2\pi i} \int_{-1}^{1} \frac{f(t)}{t-s} dt - \frac{1}{2} f(s). \tag{2.8}$$

**Proof** Formulas (2.7) and (2.8) are originally derived by Sokhotski and Plemelj for Hölder continuous functions. The existence of $\lim_{\varepsilon \to 0} \Phi(s+i\varepsilon)$ and $\lim_{\varepsilon \to 0} \Phi(s-i\varepsilon)$ for integrable functions was obtained in Privalov's series of papers. Detailed expositions and derivations of these results for integrable functions can be found from [4] and [26]. □

**Theorem 1**. *Assume that* $f(t) \in L^p(-1,1)$, $p \geq 1$ *and* $a < b$, *then* $f(t)$ *and* $F(s)$ *can be uniquely determined in* $(-1,1)$ *under either one of the following conditions:*

*1. Both $f(t)$ and $F(s)$ are known in $(a,b) \cap (-1,1) \neq \phi$, $\phi$ stands for empty set;*

*2. $F(s)$ is known in $(a,b)$ for $a>1$ or $b<-1$.*

**Proof**. Without loss of generality we assume $f(t) = 0$ in $(a,b)$. If $f(t) \neq 0$ in $(a,b)$ and $F(s)$ is known in $(a,b)$, after subtracting $\int_a^b dt f(t)/[\pi(s-t)]$ from $F(s)$, we have the desired assumption.

Let $\Phi(z)$ be the function defined by (2.6). Notice that in case 1, there is no jump in (2.7) and (2.8) because of $f(t) = 0$ in $(a,b)$, thus $\Phi(z)$ is continuous in $(a,b)$ in both cases. It follows that $\Phi(z)$ is analytical in $\{C \setminus [-1,1]\} \cup (a,b)$. Recall the fact that an analytic function on an open connected set is completely determined by its values on any set of points containing a limit point [27, 28]. It follows that $\Phi(z)$ can be uniquely determined from the values in $(a,b)$ and $f(t)$ is determined by $\Phi^+(s)-\Phi^-(s)$ in $(-1,1)$. □

**Remark 1**. Theorem 1 reveals that $f(t)$ and $F(s)$ can be obtained from their partial data while it is still lacking a numerical procedure to find them. Nonetheless, Theorem 1 yields the uniqueness of $f(t)$ and $F(s)$ from very little data. We give one example in using Theorem 1 as follows. Let

$$f_1(t) = \begin{cases} \sqrt{1-t^2} & |t| \leq 1 \\ 0 & |t| > 1 \end{cases}, \quad F_1(s) = \begin{cases} s & |s| \leq 1 \\ s - \text{sign}(s)\sqrt{s^2-1} & |s| > 1 \end{cases}. \quad (2.9)$$

Notice that $f_1(t)$ and $F_1(s)$ construct one pair of a function and its Hilbert transform. Assume that $F_1(s)$ is known on certain interval such as $(2,3)$ outside of $[-1,1]$, then for the expression of $s-\sqrt{s^2-1}$, we will construct a function $\Phi_1(z)$ such that in $C\setminus[-1,1]$, $\Phi_1(z)$ is analytical and $\Phi_1(s) = s-\sqrt{s^2-1}$ in $(2,3)$. We choose $(-\infty,-1]$ as the branch cut for $\sqrt{z+1}$ and $(-\infty,1]$ as the branch cut for $\sqrt{z-1}$, respectively. On the branch cuts, we have

$$\lim_{\varepsilon>0,\varepsilon\to 0}\sqrt{s+i\varepsilon+1} = -\lim_{\varepsilon>0,\varepsilon\to 0}\sqrt{s-i\varepsilon+1} = i\sqrt{|1+s|} \text{ for } s\in(-\infty,-1], \quad (2.10)$$

$$\lim_{\varepsilon>0,\varepsilon\to 0}\sqrt{s+i\varepsilon-1} = -\lim_{\varepsilon>0,\varepsilon\to 0}\sqrt{s-i\varepsilon-1}) = i\sqrt{1-s} \text{ for } s\in(-\infty,1]. \quad (2.11)$$

Through gluing two branch cuts together, $\sqrt{z+1}\sqrt{z-1}$ becomes analytical in $(-\infty,-1)$. It follows that $\sqrt{z+1}\sqrt{z-1}$ is analytical in $C\setminus[-1,1]$ and $\lim_{|z|\to\infty}(\sqrt{z+1}\sqrt{z-1})/z=1$. With such selection, it is straightforward to verify that $\Phi_1(z) = (z-\sqrt{z+1}\sqrt{z-1})/(2i)$ meets the second condition of Theorem 1. Then $f_1(t) = \Phi_1^+(s) - \Phi_1^-(s)$ and $F_1(s) = \Phi_1^+(s) + \Phi_1^-(s)$ in $(-1,1)$ by (2.7) and (2.8).

**Remark 2**. We point out that the condition that $f(t)$ and $F(s)$ is known on a same interval cannot be reduced in Theorem 1. We provide one counterexample if the condition of Theorem 1 is not met. For a fixed positive number $\varepsilon_0$ with $1>\varepsilon_0>0$, let $F(s)=s$ in $(-\varepsilon_0,\varepsilon_0)$, we claim that there are infinite numbers of $f_\varepsilon(t)$ such that their Hilbert transforms are equal to $s$ in $(-\varepsilon_0,\varepsilon_0)$ if $1>\varepsilon>\varepsilon_0$. Through shift and scaling, we change (2.9) to the following pair of functions

$$f_\varepsilon(t) = \begin{cases} \sqrt{\varepsilon^2-t^2} & -\varepsilon \leq t \leq \varepsilon \\ 0 & \text{otherwise}, \end{cases} \quad (2.12)$$

$$F_\varepsilon(s) = \begin{cases} s & -\varepsilon \leq s \leq \varepsilon \\ s - \text{sign}(s)\sqrt{s^2-\varepsilon^2} & \text{otherwise}. \end{cases} \quad (2.13)$$

Notice that $f_\varepsilon(t)$ and $F_\varepsilon(s)$ belong to $L_d^2(-1,1)$ and construct a pair of a function and the Hilbert transform. From the definitions of (2.12) and (2.13), it is easy to see that there are many different pairs $f_\varepsilon(t)$ and $F_\varepsilon(s)$ such that $f_\varepsilon(t)=0$ for $|t|>\varepsilon$ and $F(s)\equiv s$ in $(-\varepsilon_0,\varepsilon_0)$ are known when $(-\varepsilon_0,\varepsilon_0)$ and $(-1,1)\setminus[-\varepsilon,\varepsilon]$ is disjointed.

In SPECT, PET and other applications, the projection data in some cases can be expressed as the exponential Radon transform of certain unknown function to be solved. To deal with the truncated data as investigated in [29], one approach is to invert the cosh-weighted Hilbert transform $F_\mu(s)$, $\mu\in C$,

$$F_\mu(s) = \frac{1}{\pi}\int_{-1}^{1}\frac{\cosh[\mu(s-t)]}{s-t}f(t)dt. \quad (2.14)$$

For the pair of $f(t)$ and $F_\mu(s)$, it is easy to obtain the following corollary.

**Corollary 1**. *Let $f(t) \in L^p(-1,1)$, $p \geq 1$, then $f(t)$ and $F_\mu(s)$ can be uniquely determined in $(-1,1)$ under either one of the following conditions:*

*1. Both $f(t)$ and $F_\mu(s)$ are known in $(a,b) \cap (-1,1) \neq \phi$, $\phi$ stands for empty set;*

*2. $F_\mu(s)$ is known in $(a,b)$ for $a > 1$ or $b < -1$.*

**Proof.** We mention that $\cosh(\mu z)$ and $\sinh(\mu z)$ are entire functions. Define $\Phi_\mu(z)$ as

$$\Phi_\mu(z) = \frac{\cosh(\mu z)}{2\pi i} \int_{-1}^{1} \frac{\cosh(\mu t) f(t)}{t-z} dt - \frac{\sinh(\mu z)}{2\pi i} \int_{-1}^{1} \frac{\sinh(\mu t) f(t)}{t-z} dt. \qquad (2.15)$$

Notice that $\Phi_\mu(z)$ is analytical in $C \setminus [-1, 1]$. From the Sokhotski-Plemelj formulas, for almost every $s \in (-1,1)$, $\lim_{\varepsilon > 0, \varepsilon \to 0} \Phi_\mu(s+i\varepsilon)$ and $\lim_{\varepsilon > 0, \varepsilon \to 0} \Phi_\mu(s-i\varepsilon)$ exist and satisfy

$$\begin{aligned}
\Phi_\mu^+(s) &= \lim_{\varepsilon > 0, \varepsilon \to 0} \Phi_\mu(s+i\varepsilon) \\
&= \frac{1}{2\pi i}[\cosh(\mu s) \int_{-1}^{1} \frac{\cosh(\mu t) f(t)}{t-s} dt - \sinh(\mu s) \int_{-1}^{1} \frac{\sinh(\mu t) f(t)}{t-s} dt] \\
&\quad - \frac{1}{2}[\cosh^2(\mu s) - \sinh^2(\mu s)] f(s) \\
&= \frac{1}{2\pi i} \int_{-1}^{1} \frac{\cosh(\mu(s-t)) f(t)}{t-s} dt + \frac{1}{2} f(s),
\end{aligned} \qquad (2.16)$$

$$\begin{aligned}
\Phi_\mu^-(s) &= \lim_{\varepsilon > 0, \varepsilon \to 0} \Phi_\mu(s-i\varepsilon) \\
&= \frac{1}{2\pi i}[\cosh(\mu s) \int_{-1}^{1} \frac{\cosh(\mu t) f(t)}{t-s} dt - \sinh(\mu s) \int_{-1}^{1} \frac{\sinh(\mu t) f(t)}{t-s} dt] \\
&\quad + \frac{1}{2}[\cosh^2(\mu s) - \sinh^2(\mu s)] f(s) \\
&= \frac{1}{2\pi i} \int_{-1}^{1} \frac{\cosh(\mu(s-t)) f(t)}{t-s} dt - \frac{1}{2} f(s).
\end{aligned} \qquad (2.17)$$

Using the same arguments in the proof of Theorem 1, we prove the corollary. □

**Out-of-field-of-view in CT/SPECT scan.** To conclude this section, we apply Corollary 1 to one of the typical truncation problems in CT and SPECT when the object to be scanned is out of field of view. For SPECT, the exponential Radon transform (ERT) $p(\theta, s, \mu)$ can be obtained by multiplying the measurement with an exponential factor that is decided by the attenuation coefficient $\mu$ and the distance between the center of the rotation and the object's edge of the SPECT detector. Assume that the detector size is 2 and the object is measured within the unit circle, as shown as a thick circle in Figure 1. A typical truncation problem in SPECT can be shown in Figure 1 when the support of $f(x, y)$ is inside an ellipse whose major-axis (i.e., the longest diameter) is longer than 2 and minor-axis (i.e., the shortest diameter) is less than 2. Notice that the object is fully covered when the detector surface is parallel to x-axis while the outer part is not covered when the detector surface is parallel to y-axis.

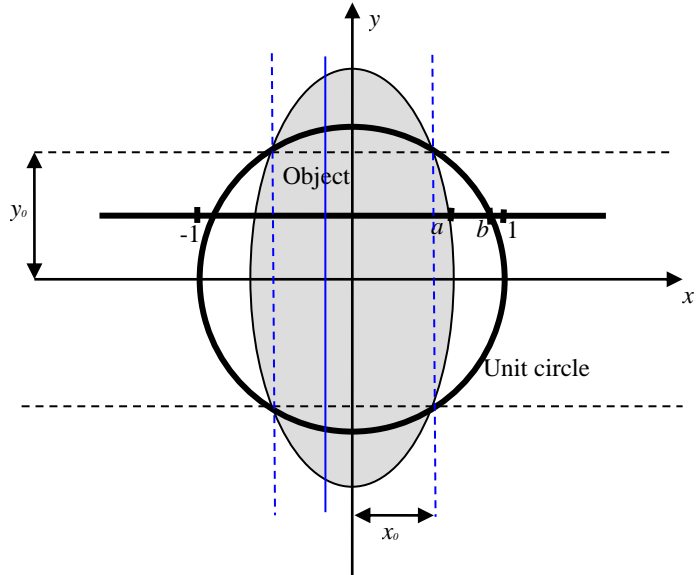

Figure 1. A SPECT imaging configuration with an elliptical object. The object is measured within the unit circle.

Let $\mu \in R^1$ and $f(x,y)$ be a function of $R^2$, the ERT is defined as

$$p(\theta, s, \mu) = \int_{-\infty}^{\infty} f(s\vec{\theta} + t\vec{\theta}^\perp) e^{\mu t} dt, \qquad (2.18)$$

where $\vec{\theta} = (\cos\theta, \sin\theta)$ and $\vec{\theta}^\perp = (-\sin\theta, \cos\theta)$. We refer to [29] for detailed data formation of $f(x,y)$ and $p(\theta, s, \mu)$ in SPECT scanning. Let $\vec{r} = (x,y)$ and define $F_y(x,y)$ as the weighted backprojection of the derivative of the ERT:

$$F_y(x,y) = -\frac{1}{2} \int_{-\pi/2}^{\pi/2} e^{-\mu \vec{r} \cdot \vec{\theta}^\perp} p'(\theta, \vec{r} \cdot \vec{\theta}) d\theta. \qquad (2.19)$$

It has been shown [9] that the weighted backprojection $F_y(x,y)$ is the cosh-weighted Hilbert transform of the object $f(x,y)$, that is,

$$F_y(x,y) = \int_{-\infty}^{\infty} \frac{\cosh(\mu\tau) f(x-\tau, y)}{\pi\tau} d\tau. \qquad (2.20)$$

Let us consider an arbitrary thick horizontal line between the two dashed lines in Fig. 1. On this thick line, $F_y(x,y)$ does not include truncation, so one is able to accurately reconstruct $f(x,y)$ by inverting an FHT defined by (2.20). There are a lot of numerical results for the inverse FHT in our paper [29]. Let $y_0$ be the distance between the dashed line and x-axis, then $f(x,y)$ is known for $|y| \le y_0$. However, $f(x,y)$ is still unknown on the points away from two dashed lines but inside the ellipse. We consider

$$F_x(x,y) = -\frac{1}{2} \int_0^{\pi} e^{-\mu \vec{r} \cdot \vec{\theta}^\perp} p'(\theta, \vec{r} \cdot \vec{\theta}) d\theta$$

$$= \int_{-\infty}^{\infty} \frac{\cosh(\mu\tau) f(x, y-\tau)}{\pi\tau} d\tau. \qquad (2.21)$$

Because of truncation, $F_x(x,y)$ is not completely available inside the entire ellipse but indeed is available if $|y| \le y_0$ on any vertical blue line between two dashed blue lines in Fig 1. Previously $f(x,y)$ has been obtained for $|y| \le y_0$. Thus on such vertical blue line, condition C1 of Corollary 1 is met, it follows that $f(x,y)$ can be uniquely determined for $|y| > y_0$. In summary, we first use the inverse FHT to reconstruct $f(x,y)$ for $|y| \le y_0$ along lines parallel to x-axis, then use the THT to uniquely determine the remaining part along lines parallel to y-axis. If $\mu = 0$, this

example is reduced to a typical X-ray CT truncation data because the bordering part of the object is out of the field of view. We summarize the preceding scanning settings to Corollary 2.

**Corollary 2** (Global uniqueness of ERT). *Let $f(x,y)$ be a continuous function with support on the ellipsoid $(\frac{x}{a})^2 + (\frac{y}{b})^2 \leq 1$. If $p(\theta,s,\mu)$ is known for $|s| \leq c$ and $\theta \in [0, 2\pi]$, here $1 \geq b > c > a > 0$ are constants. Then $f(x,y)$ is uniquely determined inside $(\frac{x}{a})^2 + (\frac{y}{b})^2 \leq 1$.*

**Proof**. First we apply Corollary 1 to (2.20), the uniqueness can be derived on the strip defined by $|y| < c$, then apply Corollary 1 to (2.21) to derive the uniqueness for $|x| < c$. It follows that the ellipsoid is fully covered.

**Corollary 3** (Local uniqueness of ERT). *Assume a continuous function $f(x,y)$ is known on the disk $|\vec{r}| \leq r_1$ and $p(\theta,s,\mu)$ is known for $|s| \leq r_2$ and $\theta \in [0, 2\pi]$, here $1 > r_1, r_2 > 0$ are two constants. Then $f(x,y)$ is uniquely determined in the regions defined by $|x| < \min(r_1, r_2)$ or $|y| < \min(r_1, r_2)$.*

**Proof**. First we apply Corollary 1 to (2.20), the uniqueness can be derived on the strip defined by $|y| < \min(r_1, r_2)$, then apply Corollary 1 to (2.21) to derive the uniqueness for $|x| < \min(r_1, r_2)$.

## III.  Chebyshev polynomial expansion of *f(t)* and *F(s)*

In this section, we only consider the THT without cosh weights. Based on the analyticity of $F(s)$, we give a formal statement on the THT problem with three conditions as follows:

**Problem**. *Find $f(t)$ in $(-1,1)$ if $f(t) \in L_a^2(-1,1)$ and $F(s)$ satisfy one of following conditions:*

  C1. $f(t)$ is known in $(a,b)$ and $F(s)$ is known in $(c,d)$ with $(a,b) \subset (c,d) \subset (-1,1)$;
  C2. $F(s)$ is known in $(c,d)$ for $d \leq -1$ or $c \geq 1$;
  C3. $f(t)$ is know in $(a,1)$ and $F(s)$ is known in $(c,d)$ for $a > -1$, $c \leq a$ and $d \geq 1$.

We point out that conditions C2 and C3 can be converted into condition C1 through shift and scaling operations, but such conversion will lose the analytical properties of $F(s)$ in different intervals. It is this reason that we use three conditions to represent the truncation problems studied in the literature. According to Theorem 1, mathematically there exists a unique solution to above truncation problem. However, to numerically find the solution remains a challenge since the analytical function $\Phi(z)$ exists but a numerical procedure is lacking to compute the solution. To the authors' knowledge, no any sort of explicit formula has been discovered to find $f(t)$ if $(c,d)$ is only a subset of $(-1,1)$. The POCS and SVD methods were suggested to estimate $f(t)$ without using a set of explicit basis functions in [13, 15-18, 21, 22]. In particular, the SVD methods in [21, 22] lack of uniqueness and are virtually impossible to lead to a numerically stable procedure since both eigenvalues and eigenfunctions do not have explicit expressions.

   In this paper, we propose two methods to numerically find the solution. The key technique is to use the Chebyshev polynomial series expansion. We express $f(t)$ and $F(s)$ in the series of Chebyshev polynomials, then convert finding $f(t)$ to estimating the coefficients through the minimization principle and an extrapolation procedure. First we introduce some results on the Chebyshev polynomials. Let $U_{-1}(s) \equiv 0$, $U_0(t) = 1$, $T_0(t) = 1$, and for $n = 1, 2, ...$, define

$$T_n(t) = \cos(n\,\mathrm{acos}(t)), \qquad (3.1)$$

$$U_n(t) = \frac{\sin((n+1)\,\mathrm{acos}(t))}{\sin(\mathrm{acos}(t))}. \qquad (3.2)$$

Here $T_n(t)$ and $U_n(t)$ are the Chebyshev polynomials of the first and second kind, respectively. Over $[-1, 1]$, we have those orthogonal relations:

$$\int_{-1}^{1} \frac{T_m(t)T_n(t)}{\sqrt{1-t^2}} dt = \begin{cases} 0 & m \neq n \\ \pi & m = n = 0 \\ \pi/2 & m = n \neq 0, \end{cases} \quad (3.3)$$

$$\int_{-1}^{1} U_m(t)U_n(t)\sqrt{1-t^2} dt = \begin{cases} 0 & m \neq n \\ \pi/2 & m = n. \end{cases} \quad (3.4)$$

From [5], the FHT of the Chebyshev polynomials on $[-1, 1]$ satisfies the following relations

$$\int_{-1}^{1} \frac{U_n(t)}{\pi(s-t)} \sqrt{1-t^2} dt = T_{n+1}(s), \quad (3.5)$$

$$\int_{-1}^{1} \frac{T_n(t)}{\pi(s-t)} \frac{1}{\sqrt{1-t^2}} dt = -U_{n-1}(s). \quad (3.6)$$

Probably the extension of (3.5-3.6) to $R^1 \setminus [-1, 1]$ was derived in the literature, but we have not seen the exact expression. Here we provide a proof based on the recurrence relations.

**Lemma 1** (Extension of Chebyshev polynomials). *For $s \in R^1 \setminus [-1, 1]$ and $n \geq 0$, we have*

$$\int_{-1}^{1} \frac{U_n(t)}{\pi(s-t)} \sqrt{1-t^2} dt = T_{n+1}(s) - U_n(s)\text{sign}(s)\sqrt{s^2-1}, \quad (3.7)$$

$$\int_{-1}^{1} \frac{T_n(t)}{\pi(s-t)} \frac{1}{\sqrt{1-t^2}} dt = T_n(s)\frac{\text{sign}(s)}{\sqrt{s^2-1}} - U_{n-1}(s). \quad (3.8)$$

**Proof**. The recurrence relations for the Chebyshev polynomials are

$$U_{n+1}(t) = 2tU_n(t) - U_{n-1}(t), \quad T_{n+1}(t) = 2tT_n(t) - T_{n-1}(t), \quad t \in R^1. \quad (3.9)$$

When $n = 0$, for $U_0(t) = 1$, equation (2.9) can be rewritten as

$$\int_{-1}^{1} \frac{U_0(t)}{\pi(s-t)} \sqrt{1-t^2} dt = T_1(s) - U_0(s)\text{sign}(s)\sqrt{s^2-1} \text{ for } s \in R^1 \setminus [-1, 1]. \quad (3.10)$$

When $n = 1$, for $U_1(t) = 2t$, we have

$$\int_{-1}^{1} \frac{U_1(t)}{\pi(s-t)} \sqrt{1-t^2} dt = 2s\int_{-1}^{1} \frac{1}{\pi(s-t)} \sqrt{1-t^2} dt + \int_{-1}^{1} \frac{2(t-s)}{\pi(s-t)} \sqrt{1-t^2} dt$$

$$= 2s\int_{-1}^{1} \frac{U_0(t)}{\pi(s-t)} \sqrt{1-t^2} dt - 1 \quad (3.11)$$

$$= 2s[T_1(s) - U_0(s)\text{sign}(s)\sqrt{s^2-1}] - T_0(s)$$

$$= T_2(s) - U_1(s)\text{sign}(s)\sqrt{s^2-1}.$$

When $n \geq 1$, using the orthogonal relations in (3.4) we have

$$\int_{-1}^{1} \frac{U_{n+1}(t)}{\pi(s-t)} \sqrt{1-t^2} dt = \int_{-1}^{1} \frac{2tU_n(t) - U_{n-1}(t)}{\pi(s-t)} \sqrt{1-t^2} dt$$

$$= 2s\int_{-1}^{1} \frac{U_n(t)}{\pi(s-t)} \sqrt{1-t^2} dt - \frac{2}{\pi}\int_{-1}^{1} U_n(t)\sqrt{1-t^2} dt - \int_{-1}^{1} \frac{U_{n-1}(t)}{\pi(s-t)} \sqrt{1-t^2} dt \quad (3.12)$$

$$= 2s\int_{-1}^{1} \frac{U_n(t)}{\pi(s-t)} \sqrt{1-t^2} dt - \int_{-1}^{1} \frac{U_{n-1}(t)}{\pi(s-t)} \sqrt{1-t^2} dt.$$

Combining equations (2.9, 3.10-3.12), we have proven (3.7). Similar arguments lead to (3.8). □

A recent comprehensive review on the Chebyshev polynomials can be found in [25]. We collect several known results in Lemma 2 for the use in this paper.

**Lemma 2** (Properties of Chebyshev polynomials). *For $t \in [-1, 1]$, $s \in R^1$ and $n \geq 0$, we define*

$$\tilde{U}_n(t) = \sqrt{1-t^2}\, U_n(t), \quad (3.13)$$

$$\tilde{T}_n(s) = \begin{cases} T_n(s) & |s| \leq 1 \\ T_n(s) - U_{n-1}(s)\,\text{sign}(s)\sqrt{s^2-1} & |s| > 1. \end{cases} \quad (3.14)$$

*Then we have four properties:*

1. $\{T_n(t), n \geq 0\}$ and $\{\tilde{U}_n(t), n \geq 0\}$ *construct a complete orthogonal basis set of* $L^2_d(-1,1)$;
2. $\{T_n(t), n \geq 1\}$ and $\{\tilde{U}_n(t), n \geq 1\}$ *are the complete orthogonal basis set of* $Ł^2_d(-1,1)$;
3. $\{\tilde{U}_n(t), \tilde{T}_n(s)\}$, $n \geq 0$, *satisfies* $\|\tilde{U}_n(t)\|_{L^2(-1,1)} = \|\tilde{T}_n(s)\|_{L^2(R^1)}$;
4. $\{\tilde{U}_n(t), \tilde{T}_n(s)\}$ *forms an SVD for the Hilbert transform from* $L^2_d(-1,1)$ *to its range.*

Assume that $f(t)$ and $F(s)$ belong to $L^2_d(-1,1)$, according to Lemma 2, we can express the pair of $f(t)$ and $F(s)$ in the following series

$$f(t) = \sqrt{1-t^2}\sum_{n\geq 1}^{\infty} c_n U_{n-1}(t),\quad t \in (-1,1), \quad (3.15)$$

$$F(s) = \sum_{n\geq 1}^{\infty} c_n \tilde{T}_n(s),\quad s \in R^1. \quad (3.16)$$

In the sense of approximation, on $[-1,1]$, we may have the following relations

$$f(t) \approx \sqrt{1-t^2}\sum_{n\geq 1}^{N} c_n U_{n-1}(t), \quad (3.17)$$

$$F(s) \approx \sum_{n\geq 1}^{N} c_n \tilde{T}_n(s). \quad (3.18)$$

For the polynomial interpolation, we introduce the following coordinate transformation

$$u = v - \sqrt{v^2-1},\quad v = \frac{u^2+1}{2u}\quad \text{for } v \in (1,\infty) \text{ and } u \in (0,1). \quad (3.19)$$

**Theorem 2.** *Let* $f(t)$ *and* $F(s)$ *meet one of C1, C2 and C3. If* $f(t) \in L^2_d(-1,1)$ *and can be expressed as a finite series*

$$f(t) = \sqrt{1-t^2}\sum_{n\geq 1}^{N} c_n U_{n-1}(t), \quad (3.20)$$

*where* $\{c_n\}$ *are unknown coefficients, then there exists an explicit procedure to find* $\{c_n\}$.

**Proof.** With the assumption of (3.20), $F(s)$ can be expressed as

$$F(s) = \sum_{n\geq 1}^{N} c_n \tilde{T}_n(s). \quad (3.21)$$

We first prove the theorem for condition C1. In $(c,d)$, $F(s)$ is a polynomial with the following expression

$$F(s) = \sum_{n\geq 0}^{N} a_n s^n. \quad (3.22)$$

Select $N+1$ distinct sampling points $\{s_m, 0 \leq m \leq N\}$ in $(c,d)$, and let $P(s)$ be the Lagrange interpolation polynomial on these points

$$P(s) = \sum_{n=0}^{N}\Big[\prod_{\substack{0\leq m\leq N\\ m\neq n}}\frac{s-s_m}{s_n-s_m}F(s_n)\Big]. \quad (3.23)$$

Rewriting (3.23) to (3.22), we obtain the coefficients $\{a_n\}$. It follows that $\{c_n\}$ can be obtained by expressing $s^n$ in the Chebyshev polynomial series.

For condition C2, without loss of generality, we assume $c \geq 1$, then in $(c,d)$, $F(s)$ is

$$F(s) = \sum_{n\geq 1}^{N} c_n[T_n(s) - U_{n-1}(s)\sqrt{s^2-1}] = \sum_{n\geq 1}^{N} c_n(s - \sqrt{s^2-1})^n. \quad (3.24)$$

Here we used the following identity
$$T_n(s) - U_{n-1}(s)\text{sign}(s)\sqrt{s^2-1} = (s - \text{sign}(s)\sqrt{s^2-1})^n. \tag{3.25}$$

By coordinate transformation (3.19), in $(d - \sqrt{d^2-1}, c - \sqrt{c^2-1})$, (3.24) becomes a polynomial
$$F(u) = \sum_{n\geq 1}^{N} c_n u^n. \tag{3.26}$$

Selecting $N+1$ distinct sampling points in $(d - \sqrt{d^2-1}, c - \sqrt{c^2-1})$, and then by the Lagrange interpolation polynomial, we can obtain $\{c_n\}$. The condition *C3* can be converted into either C1 or C2. This completes the proof. □

The Lagrange interpolations are notoriously ill-posed for high-order polynomials. In the finite dimension, the explicit procedure to find $f(t)$ may be only meaningful in the perfect world. In practical applications such as CT/SPECT imaging, $F(s)$ may contain measurement and computation errors and will be approximated by (3.17). In order to fully use the available data from both $f(t)$ and $F(s)$, we suggest a minimization criterion to estimate the coefficients for all three conditions of C1, C2 and C3.

**Minimization criterion**. On $[-1, 1]$, the Chebyshev polynomials have the orthogonal relations (3.3) and (3.4). Outside of $[-1, 1]$, we only have a global Plancherel formula for $\{\tilde{U}_n(t), \tilde{T}_n(s)\}$ in $L^2$ space as described in Lemma 2. Assume $f(t)$ in known in $(a,b) \subset (-1,1)$ and $F(s)$ in known in $(c,d) \subset R^1$, we use the least square criterion to define a cost function $M(\{c_n\})$ as follows

$$\begin{aligned}M(\{c_n\}) = &\int_{(a,b)} |f(t) - \sqrt{1-t^2}\sum_{n\geq 1}^{N} c_n U_{n-1}(t)|^2 |d\arccos(t)| \\ &+ \int_{(c,d)\cap(-1,1)} |F(s) - \sum_{n\geq 1}^{N} c_n T_n(s)|^2 |d\arccos(s)| \\ &+ \int_{(c,d)\setminus[-1,1]} |F(s) - \sum_{n\geq 1}^{N} c_n (s - \text{sign}(s)\sqrt{s^2-1})^n|^2 \, ds.\end{aligned} \tag{3.27}$$

Here we point out that the last term of (3.27) outside of $[-1, 1]$ takes polynomials whose term is contracting due to (3.25). Also we use the Stieltjes type integral through $d\arccos(t)$ to avoid the singularity at the end points of $[-1, 1]$. On $[-1, 1]$ the $L^2_d(-1, 1)$ norm is used while the $L^2$ norm is adopted outside of $[-1, 1]$. Equation (3.27) is one way to define a cost function. There may exist other optimized ways to define the cost function concerning *a prior* information.

Assume that all the sampling points in $(a, b)$ and $(c, d)$ are equally spaced with the same interval $\Delta$. Let $\{t_k, 1 \leq k \leq K\}$ be the available sampling points on $(a, b)$, $\{s_i, 1 \leq i \leq I\}$ be the available sampling points in $(c, d) \cap (-1, 1)$, and $\{s_j, 1 \leq j \leq J\}$ be the available sampling points in $(c, d) \setminus [-1, 1]$. In a form of simple numerical integrals, the cost function $M(\{c_n\})$ in (3.27) may be approximated by

$$\begin{aligned}M(\{c_n\}) \approx &\sum_{k=1}^{K} |f(t_k) - \sqrt{1-t_k^2}\sum_{n\geq 1}^{N} c_n U_{n-1}(t_k)|^2 |\arccos(t_k + 0.5\Delta) - \arccos(t_k - 0.5\Delta)| \\ &+ \sum_{i=1}^{I} |F(s_i) - \sum_{n\geq 1}^{N} c_n T_n(s_i)|^2 |\arccos(s_i + 0.5\Delta) - \arccos(s_i - 0.5\Delta)| \\ &+ \sum_{j=1}^{J} |F(s_j) - \sum_{n\geq 1}^{N} c_n (s_j - \text{sign}(s_j)\sqrt{s_j^2-1})^n|^2 \, \Delta.\end{aligned} \tag{3.28}$$

Looking for $\{c_n\}$ to minimize $M(\{c_n\})$ leads to an estimate of $f(t)$ in the expression of (3.17). Cost function (3.28) can be used to find the solution in the intervals described in [30]. Numerical algorithms in [31] can be used to find $\{c_n\}$ to minimize (3.28). Let $\{\hat{c}_n\}$ be one set of coefficients to reach the minimum of $M(\{c_n\})$, then the estimated solution is

$$\hat{f}(t) = \sqrt{1-t^2} \sum_{n\geq 1}^{N} \hat{c}_n U_{n-1}(t). \tag{3.29}$$

In $L_d^2(-1,1)$, the error between estimated $\hat{f}(t)$ and the original function $f(t)$ is

$$\int_{-1}^{1} \frac{|\hat{f}(t) - f(t)|^2}{\sqrt{1-t^2}} dt = \sum_{n=1}^{N} |\hat{c}_n - c_n|^2 + \sum_{n=N+1}^{\infty} |c_n|^2. \tag{3.30}$$

**Extrapolation procedure**. Condition C1 is one of the typical truncation problems arising in CT/SPECT as previously described in Section II for the out-of-field-of-view in CT/SPECT scanning. Motivated by the iterative schema for the band-limited signal extrapolation in [32, 33], we construct an iterative procedure to find $f(t)$ under condition C1. We denote by $\mathbf{H}$ the finite Hilbert transform (1.0), and let $\mathbf{H}^{-1}$ be the following inversion formula

$$f(t) = \frac{1}{\pi} \int_{-1}^{1} \frac{F(s)}{s-t} \frac{\sqrt{1-t^2}}{\sqrt{1-s^2}} ds. \tag{3.31}$$

Here we assume $(a,b) \subset (-1,1)$, $(c,d) \subset (-1,1)$ and $(a,b) \cap (c,d) \neq \phi$. Define an initial pair as

$$F^{(0)}(s) = \begin{cases} F(s) & s \in (c,d) \\ a\ guess & Otherwise \end{cases}, \tag{3.32}$$

$$f^{(0)}(t_n) = \begin{cases} f(t) & t \in (a,b) \\ [\mathbf{H}^{-1} F^{(0)}](t) & Otherwise \end{cases}. \tag{3.33}$$

Then we suggest the following iterative procedure

$$F^{(k+1)}(s) = \begin{cases} F(s) & s \in (c,d) \\ [\mathbf{H} f^{(k)}](s) & Otherwise \end{cases}, \tag{3.34}$$

$$f^{(k+1)}(t_n) = \begin{cases} f(t) & t \in (a,b) \\ [\mathbf{H}^{-1} F^{(k+1)}](t) & Otherwise \end{cases}. \tag{3.35}$$

For the band-limited signal extrapolation, the operator $\mathbf{H}$ is the Fourier transform [32, 33]. In [34], the author found one minor issue of [33] in using that schema. This iteration also falls in the concept of the POCS. Using the Plancherel formula in (2.5), if $f^{(k)}(t) \neq f(t)$, we have the following strictly decreasing relations

$$\begin{aligned}
\left\| f^{(k+1)}(t) - f(t) \right\|_{L_w} &\leq \left\| [\mathbf{H}^{-1} F^{(k+1)}](t) - f(t) \right\|_{L_d} \\
&= \left\| F^{(k+1)}(s) - F(s) \right\|_{L_d} \\
&< \left\| \mathbf{H} f^{(k)}(t) - F(s) \right\|_{L_d} \\
&= \left\| f^{(k)}(t) - f(t) \right\|_{L_d}.
\end{aligned} \tag{3.36}$$

However we are not able to prove the convergence in $L_d^2(-1,1)$. In order to use the fast cosine and sine transformations of [31, 35] to calculate the Hilbert transform and its inversion, for $m, n = 0, \cdots, N-1$, we choose the so-called *Chebyshev-Gauss-Lobatto (CGL)* sampling points as

$$t_m = \cos(\frac{m}{N}\pi), \quad s_m = \cos(\frac{m+0.5}{N}\pi). \tag{3.37}$$

According to the trigonometric expression of the FHT and its inversion in [24], on the *CGL* sampling points, we have

$$f(t_m) \approx \sum_{n=0}^{N-1} c_n \sin(\frac{m}{N} n\pi), \quad (3.38)$$

$$F(s_m) \approx \sum_{n=0}^{N-1} c_n \cos(\frac{m+0.5}{N} n\pi). \quad (3.39)$$

Condition C1 is equivalent to that $F(s_m)$ is known for $m_1 \leq m \leq m_2$ and $f(t_m)$ is known for $m_3 \leq m \leq m_3$. Then the task is to estimate $f(t_m)$ for $m < m_3$ and $m > m_4$. We denote by $\mathbf{M}_1$ and $\mathbf{M}_2$ the discrete versions of $\mathbf{H}$ and $\mathbf{H}^{-1}$, respectively. As shown in [24, 25, 31], $\mathbf{M}_1$ and $\mathbf{M}_2$ can be expressed as the combination of sine and cosine transforms. Then the discrete versions of (3.34-3.35) become

$$F^{(0)}(s_m) = \begin{cases} F(s_m) & m_1 \leq m \leq m_2 \\ a\ guess & otherwise, \end{cases} \quad (3.40)$$

$$f^{(0)}(t_m) = \begin{cases} f(t_m) & m_3 \leq m \leq m_4 \\ \{\mathbf{M}_2 \vec{F}^{(0)}\}_m & otherwise, \end{cases} \quad (3.41)$$

$$F^{(k+1)}(s_m) = \begin{cases} F(s_m) & m_1 \leq m \leq m_2 \\ \{\mathbf{M}_1 \vec{f}^{(k)}\}_m & otherwise, \end{cases} \quad (3.42)$$

$$f^{(k+1)}(t_m) = \begin{cases} f(t_m) & m_3 \leq m \leq m_4 \\ \{\mathbf{M}_2 \vec{F}^{(k)}\}_m & otherwise. \end{cases} \quad (3.43)$$

If $\{f^{(k)}(t_m)\} \neq \{f(t_m)\}$ then

$$\|\{f^{(k+1)}(t_m)\} - \{f(t_m)\}\|_2 < \|\{f^{(k)}(t_m)\} - \{f(t_m)\}\|_2. \quad (3.44)$$

It follows that the sequence of vectors $\{f^{(k)}(t_m)\}$ converges by the fact that a bounded set in finite dimension of Euclid space is compact.

## IV. Computer simulation results

As pointed out in the beginning of Section III, conditions C2 and C3 can be always converted to condition C1, and C1 is the common truncation problem arising in CT/SPECT scanning. Also the extrapolation procedure for condition C1 is very simple to numerically realize. In this paper, we will perform the computer simulations for the iterative procedure of (3.42, 3.43) for condition C1. Through scaling and shift on (2.9), we construct the following pair of functions

$$f(t) = \begin{cases} \sqrt{0.64 - (t+0.1)^2} & -0.9 \leq t \leq 0.7 \\ 0 & otherwise, \end{cases} \quad (4.1)$$

$$F(s) = \begin{cases} s+0.1 & -0.9 \leq s \leq 0.7 \\ s+0.1 - \text{sign}(s+0.1)\sqrt{(s+0.1)^2 - 0.64} & otherwise. \end{cases} \quad (4.2)$$

Restricted on $[-1,1]$, $f(t)$ and $F(s)$ construct a pair of a function and the Hilbert transform. We will use (4.1) and (4.2) to verify the iterative procedure of (3.42, 3.43). The *CGL* sampling grid of $\{s_m\}$ and $\{t_m\}$ of (3.37) will be used in the computer simulations. The number of sampling points is $N=256$ in all numerical simulations. Since $\{t_m\}$ is not evenly sampled on $[-1,1]$, for display purpose, we will resample $f(t)$ on $\{t_k\}$, here $t_k = (2k+1-256)/256$, $k=0,\cdots,255$. The resampling method is to first find the coefficients $\{c_n\}$ and then use the following recurrence formulas

$$U_{n+1}(t_k) = 2t_k U_n(t_k) - U_{n-1}(t_k), \quad f(t_k) = \sqrt{1-t_k^2} \sum_{n=0}^{255} c_n U_{n-1}(t_k). \quad (4.3)$$

Similarly, we will use $\{s_k\}$, $s_k = (2k+1-256)/256$, $k = 0, \cdots, 255$, for display of $F(s)$ through the following recurrence formulas

$$T_{n+1}(s_k) = 2s_k T_n(s_k) - T_{n-1}(s_k), \quad F(s_k) = \sum_{n=0}^{255} c_n T_n(s_k). \tag{4.4}$$

Thus, in all numerical simulations, we use the *CGL* sampling grid $\{s_m\}$ and $\{t_m\}$ for calculations but we take the evenly sampled grid $\{t_k\}$ and $\{s_k\}$ for display. We show $f(t)$ and $F(s)$ below.

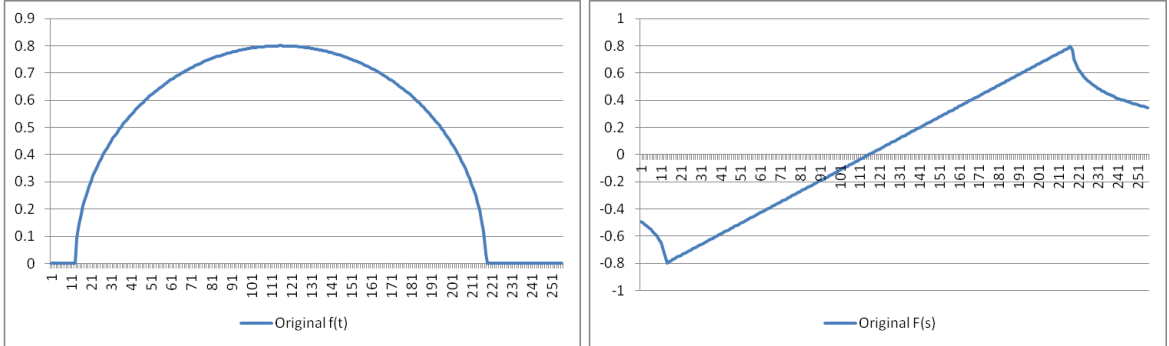

Fig 1. Left: original $f(t)$ on $\{t_k\}$; Right: $F(s)$ on $\{s_k\}$.

In the numerical experiemnts, $F(s_m)$ with $32 \leq m < 224$ and $f(t_m)$ with $64 \leq m < 192$ will be used as the initial data in (3.40, 3.41). Then we repeat 30 iterations of (3.42, 3.43) to extrapolate $F(s_m)$ and $f(t_m)$ in the outer region. The extrapolated $F(s)$ and $f(t)$ are shown in Figs. 2 and 3.

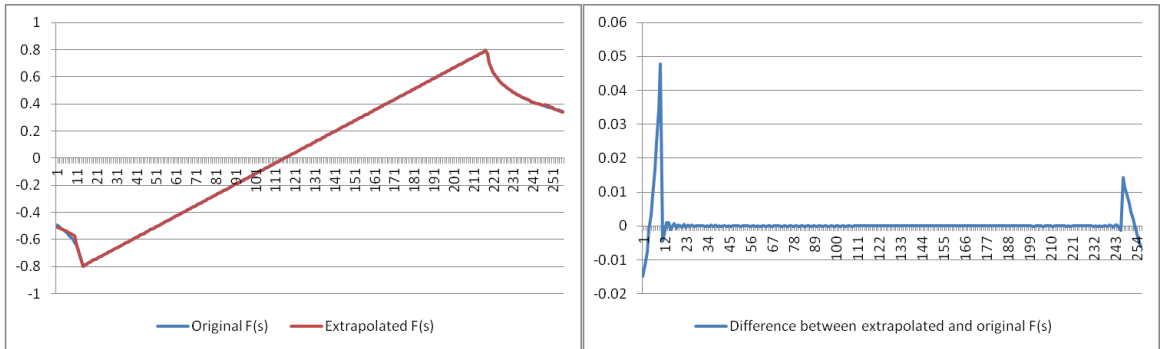

Fig 2. Original and extrapolated $F(s)$ on $\{s_k\}$.

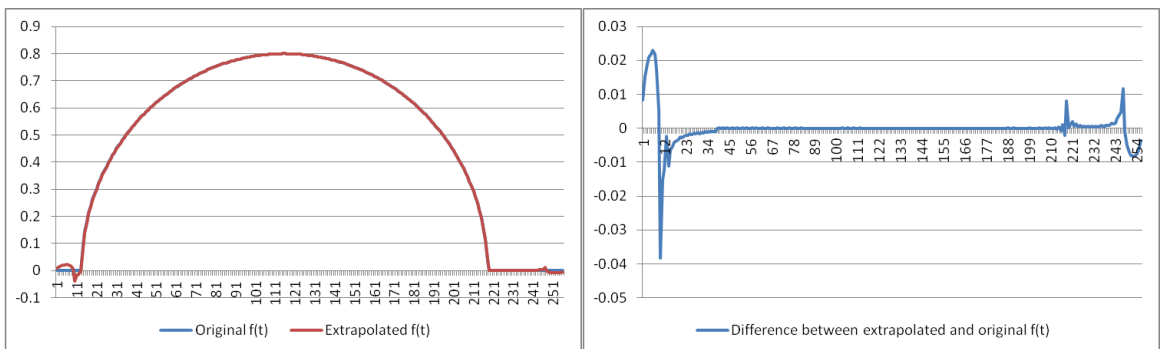

Fig 3. Original and extrapolated $f(t)$ on $\{t_k\}$.

We mention that the above truncation data setting is the condition used in [22]. As shown in Figs 2 and 3, the iterative procedure (3.42, 3.43) does extrapolate both $f(t)$ and $F(s)$ to the outer region in an accurate manner. Even under moderate level of noise, the extrapolation procedure still works well as shown in the Figs 4 and 5.

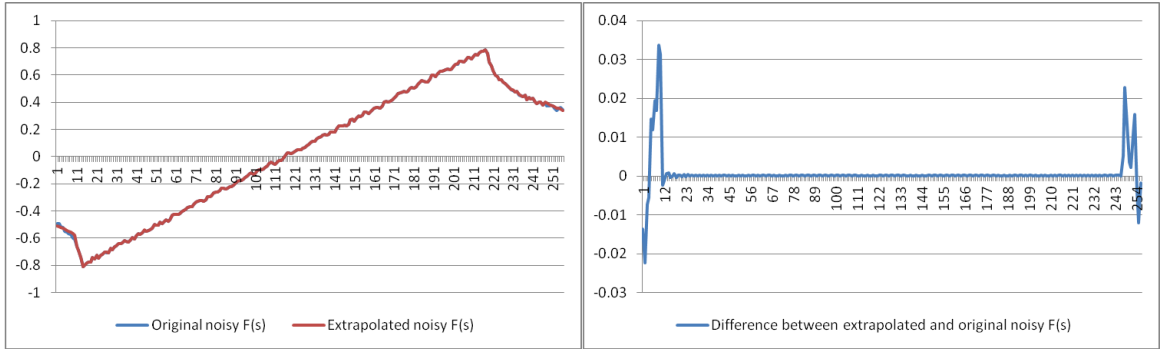

Fig 4. Original and extrapolated noisy $F(s)$ on $\{s_k\}$.

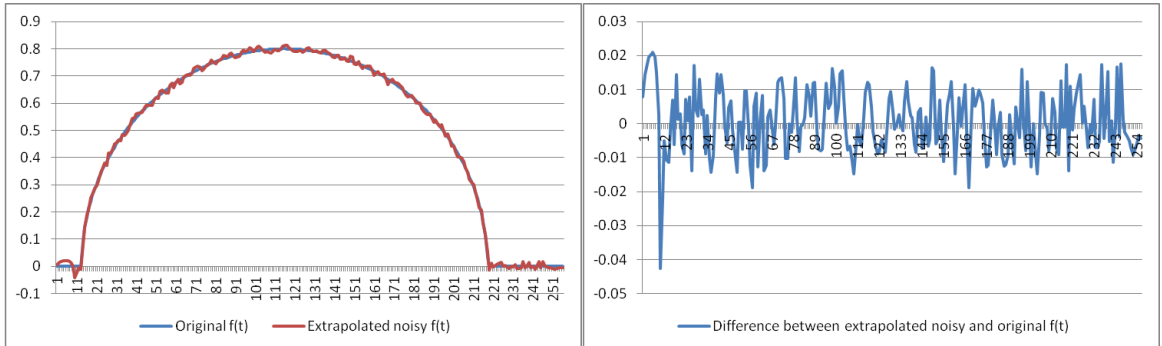

Fig 5. Original and extrapolated noisy $f(t)$ on $\{t_k\}$.

In conclusion, the iterative procedure (3.42, 3.43) is easy to carry out and seems to provide a reasonable solution for the truncation problem under condition C1.

## V. Discussion and Conclusion

In summary, we have obtained a stronger uniqueness result under weaker conditions compared with the existing works [13, 15-18]. The arguments used in [13, 15-18] cannot handle the uniqueness in the context of exponential Radon transform. Because of the powerful Sokhotski–Plemelj formulas, our uniqueness result can be applicable to the exponential Radon transform. As an application of our uniqueness result, the solution by the SVD schema in [21, 22] can be unique under the conditions used in [13, 15-18]. Following the early idea in [5], we have found that the Chebyshev polynomials can be used to construct an SVD for the truncated Hilbert transform from functions of $[-1, 1]$ to functions of $R^1$. The Chebyshev polynomials have explicit expressions and can be computed using fast algorithms such as sine and cosine transforms [31, 35, 36]. For condition C1, the implementation of iterative procedure (3.42, 3.43) is very easy and the computer simulation results are very promising. From the numerical realization standpoint, using the Chebyshev polynomial series expansion has an advantage over other existing methods.

The SVD schema in [21, 22] was obtained from $L^2(-1,1)$ to a subspace. As pointed out in [23], for $F(s) = 1$ in $(-1,1)$, its inverse FHT $-t/\sqrt{1-t^2}$ does not belong to $L^2(-1,1)$. This implies that the SVD scheme in [22] may not produce a convergent series. In this paper, we consider $f(t)$ and $F(s)$ in $L_d^2(-1,1)$ in order to use the injectivity of the FHT. Furthermore, with the Chebyshev polynomial series expansions (3.15) and (3.16), a prior knowledge of $f(t)$ can be considered in both minimization criterion (3.27) and the iterative procedure (3.42, 3.43).

The minimization criterion is more general and can be applicable to the THT data settings considered in [13, 15-18, 21, 22] through selecting the location of $[a, b]$, for example the case in [13] is the interval with $a \in (-1, 1)$ and $b \in (1, \infty)$, the case in [15-18, 22] is the interval with $a, b \in (-1, 1)$, and the case in [21] is the interval with $a, b \in (1, \infty)$. However, finding the minimum point of (3.28) may not be numerically easy as the scheme (3.42, 3.43). Theorem 2 may be not very useful to the numerical realization, but it can be used to obtain an initial estimate of $\{c_n\}$ by using the low-order polynomial interpolation toward finding the final solution of $\{c_n\}$ to reach the minimum of $M(\{c_n\})$ and providing a reasonable initial guess of $F(s)$. Due to the evaluation of high-order polynomials, to numerically find the minimum point of (3.28) is not trivial. For example, the double type only has 15~16 exact digits, then for $n \geq 15$, $|s| \leq 0.1$, the coefficients of $s^n$ may not be accurate because of round-off error. We used equally spaced sampling points in (3.28) to show a simple discrete version of (3.27), however this type of sampling causes the *Runge Phenomenon* near the boundary inside $[-1, 1]$. We mention that [36] includes a graphic tutorial on the numerical behavior of the Chebyshev polynomials. It may be interesting to study the numerical behaviors of the minimization criterion compared with the extrapolative procedure (3.42, 3.43).

To conclude the paper, we like to mention two papers [37, 38] on the explicit inversion formulas of cosh-weighted Hilbert transform in [37] and numerical approximation in [38]. Currently we are investigating the numerical characteristics of the inversion formulas of [37].